\documentclass[numbers, square]{article}
\usepackage{arxiv}

\usepackage{natbib}

\usepackage{authblk}
\usepackage[utf8]{inputenc} 
\usepackage[T1]{fontenc}    

\usepackage{graphicx}
\usepackage[nobottomtitles]{titlesec}

\usepackage{amssymb,amsmath,amsfonts,amsthm}

\usepackage[nottoc]{tocbibind}

\usepackage{hyperref}       

\author[1,2]{\small Baptiste Barreau}
\author[2]{\small Laurent Carlier}
\date{}

\affil[1]{\footnotesize Université Paris-Saclay, CentraleSupélec, Mathématiques et Informatique pour la Complexité et les Systèmes, 3 rue Joliot-Curie, 91192 Gif-sur-Yvette, France}
\affil[2]{\footnotesize BNP Paribas Corporate and Institutional Banking, Global Markets Data \& Artificial Intelligence Lab, 20 boulevard des Italiens, 75009 Paris, France}
\affil[ ]{\texttt{\{baptiste.barreau, laurent.carlier\}@bnpparibas.com}}

\begin{document}

\title{History-Augmented Collaborative Filtering for Financial Recommendations}
\maketitle

\begin{abstract}
In many businesses, and particularly in finance, the behavior of a client might drastically change over time. It is consequently crucial for recommender systems used in such environments to be able to adapt to these changes. In this study, we propose a novel collaborative filtering algorithm that captures the temporal context of a user-item interaction through the users' and items' recent interaction histories to provide dynamic recommendations. The algorithm, designed with issues specific to the financial world in mind, uses a custom neural network architecture that tackles the non-stationarity of users' and items' behaviors. The performance and properties of the algorithm are monitored in a series of experiments on a G10 bond request for quotation proprietary database from BNP Paribas Corporate and Institutional Banking.
\end{abstract}


\section{Introduction}
\label{intro}

In a financial market, liquidity is provided by \textit{market makers}, whose role is to constantly offer both buy (\textit{bid}) and sell (\textit{ask}) prices. Market makers benefit from the difference between the two, called the \textit{bid-ask spread}. Corporate and institutional banks such as BNP Paribas CIB play the role of market makers in financial markets across many asset classes and their derivatives. When a client wants to trade a financial product, she either requests prices on an electronic platform where many different market makers operate in a process called a \textit{request for quotation} (RFQ), or contact a salesperson of a bank. Reciprocally, salespeople can also directly contact clients and suggest relevant trade ideas to them, e.g., financial products held by the bank and on which it might offer a better price than its competitors. Proactive salespeople, which are particularly important for the bank, help manage financial inventories to minimize the risk to which the bank is exposed, and serve better their clients when correctly anticipating their needs. Providing salespeople with an RFQ recommender system would support their proactivity by allowing them to navigate the complexity of the markets more easily. Our goal is to design a financial recommender system that suits the particularities of the financial world to assist salespeople in their daily tasks. RFQ recommendation is an implicit feedback problem, as we do not explicitly observe clients' opinions about the products they request. Implicit feedback is a classic recommender system setup already addressed by the research community, e.g., in \citep{hu2008collaborative}. The financial environment, however, brings about specific issues that require attention. To that end, the algorithm we introduce here has three main aims: 

\begin{itemize}
    \item \textbf{To incorporate time.} In a classic e-commerce environment, leaving aside popularity and seasonal effects, recommendations provided at a given date may remain relevant for a couple of months, since users' shopping tastes do not markedly evolve with time. In the financial world, a user is interested in a financial product at a given date not only because of the product's intrinsic characteristics but also because of its current market conditions. Time is consequently crucial for RFQ prediction and should be taken into account in a manner that allows for future predictions.
    \item \textbf{To obtain dynamic embeddings.} Being able to capture how clients (resp. financial products) relate to each other and how this relationship evolves with time is of great interest for business, as it allows getting a deeper understanding of the market. One of our goals is consequently to keep the global architecture of matrix factorization algorithms, where the recommendation score of a \textit{(client, product)} couple is given by the scalar product of their latent representations.
    \item \textbf{To symmetrize users and items.} Classic recommender systems focus on the user-side. However, the product-side is also of interest in the context of a corporate bank. To control her risk, a market maker needs to control her positions, i.e., make sure she does not hold a given position for too long. Consequently, coming up with a relevant client for a given asset is equally important, and the \textit{symmetry of clients and assets} will be integrated here in both our algorithms and evaluation strategies.
\end{itemize}

We tackle these goals with a context-aware recommender system that only uses client-product interactions as its signal source. In this work, the context is dynamic, and is inferred at a given time from the previous interactions of clients and products. The terms clients and users (resp. financial products/assets and items) will be used interchangeably to match the vocabulary used in recommender systems literature.

\section{Related work}

The work presented in this article is a context-aware and time-dependent recommender system, which are both active areas of research \citep{adomavicius2011context, shi2014collaborative}. Notably, in \citep{hariri2014context}, the recommendations of a bandit algorithm are dynamically adapted to contextual changes. In \citep{aghdam2015adapting}, the recommendations of a hierarchical hidden Markov model are contextualized based on the users' feedback sequences. Down-weighing past samples in memory-based collaborative filtering algorithms helps better match temporal dynamics \citep{ding2005time}. The latent factors of a matrix factorization can also directly include temporal dynamics \citep{koren2009collaborative}. Temporal probabilistic matrix factorization \citep{chenyi2014latent} introduces dynamics of users' latent factors with a time-invariant transition matrix that can be computed using Bayesian approaches, thereby extending probabilistic matrix factorization \citep{mnih2008probabilistic}. It is also possible to enhance latent factor models with Kalman filters \citep{sarkar2007latent} or recurrent neural networks \citep{wu2017recurrent} to capture the dynamics of rating vectors.

Tensor factorization models handle time by considering the user-item co-occurrence matrix as a three-dimensional tensor where the additional dimension corresponds to time \citep{xiong2010temporal, wu2018neural}, and where temporal dynamics can be accounted for using recurrent neural networks \citep{wu2018neural} --- these approaches, however, do not allow for future predictions. Using historical data to introduce dynamics in a neural network recommender system was done in \citep{covington2016deep}, where users are assimilated to their items' histories. The Caser algorithm \citep{tang2018perso} uses convolutional filters to embed histories of previous items and provide dynamic recommendations to users. 
 
The algorithm introduced in this work is to some extent reminiscent of graph neural networks \citep{hamilton2017inductive} and their adaptation to recommendation \citep{wang2019neural}. Using time to enhance recommendations with random walks on bipartite graphs was explored in \citep{xiang2010temporal}. How graph neural networks behave with dynamic bipartite graphs is to the best of our knowledge yet to be discovered and could lead to an extension of this work. A first attempt at financial products recommendation was made in \citep{wright2018machine} with the particular example of corporate bonds. 

\section{Methodology}
\label{methodo}

We introduce a neural network architecture that aims at producing recommendations through dynamic embeddings of users and items, that we call \textit{History-augmented Collaborative Filtering} (HCF). 

Let $U$ be the set of all users and $I$ the set of all items. Let us note $x_{u} \in \mathbb{R}^{d}$ for $u \in U$, $x_{i} \in \mathbb{R}^{d}$ for $i \in I$ the static $d$-dimensional embeddings of all the users and items we consider. Let $t \in [\![ 0~; \infty]\!]$ be a discrete time-step, taken as days in this work. For a given $u$, at a given $t$, we define $\mathbf{h^{t}_{u}}$ as the items' history of $u$, i.e., the set of items found to be of interest to $u$ in the past --- we respectively define $\mathbf{h^{t}_{i}}$ as item $i$ users' history. We use here as histories the last $n$ events that happened strictly before $t$ to either user $u$ or item $i$. If at a given $t$ we observe $0 \leq n' < n$ previous events, histories are only formed of those $n'$ events. Users are consequently on a same event scale --- high-activity users' histories may span a couple of days, whereas low-activity ones may span a couple of months (resp. for items). Histories formed of items/users of interest in a past window of fixed size were also tried, but led to inferior performance.
  
\begin{figure}[h]
    \centerline{\includegraphics[width=0.8\linewidth]{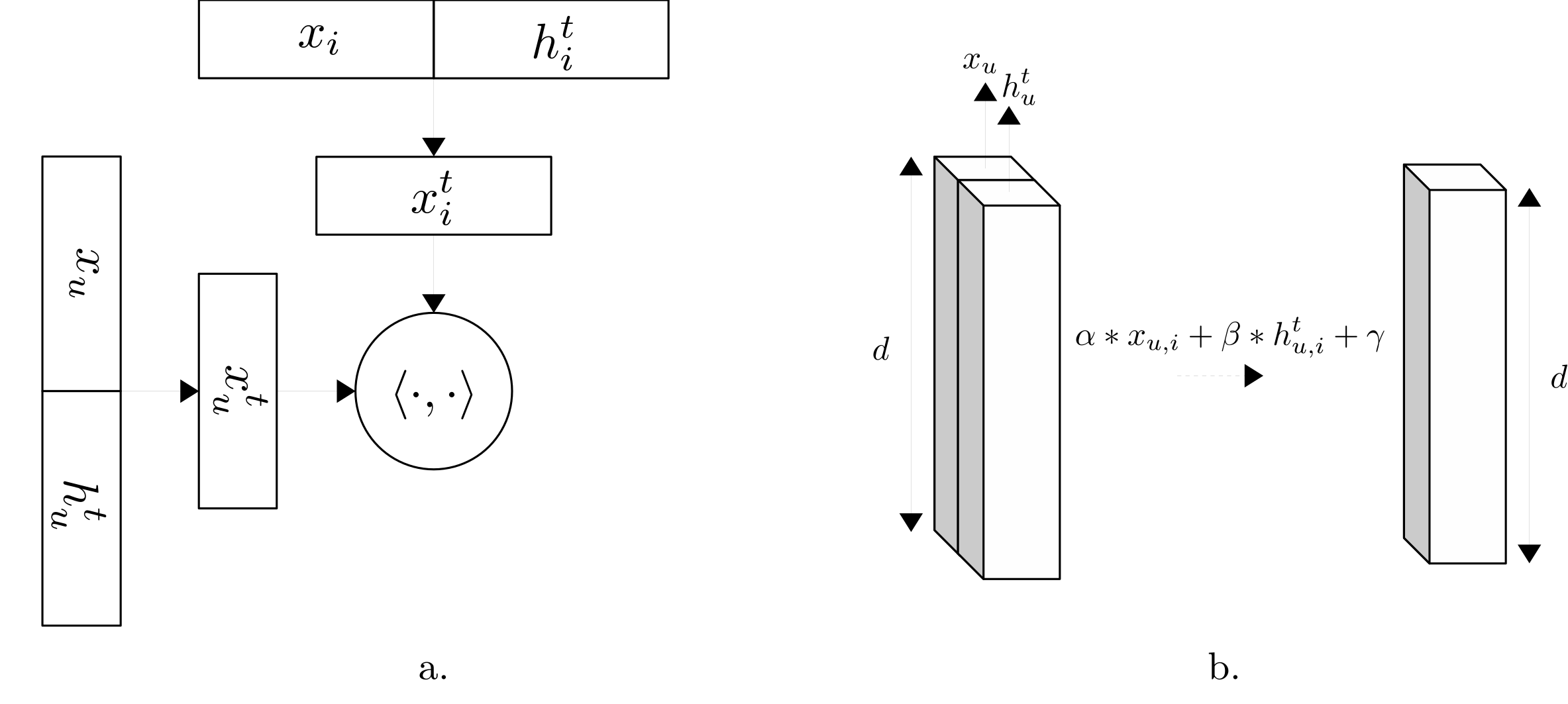}}
    \caption{\textbf{a.} Global architecture of the HCF network, composed of two symmetric user and item blocks. \textbf{b.} Illustration of the application of a one-dimensional convolution filter of kernel size $1$ along the embedding dimension axis of our inputs $x_{u}$ and $h^{t}_{u}$ for a user $u \in U$, where $\alpha, \beta, \gamma$ are the parameters of the convolution filter and $l \in [\![1; d ]\!]$. The same computation holds for items $i \in I$ respectively.}
    \label{archi}
\end{figure}

Figure \ref{archi}.a shows the global architecture of the HCF network. It is composed of two symmetric blocks --- a \textit{user block} and an \textit{item block}. At a given time $t$, the user block of HCF produces dynamic user embeddings $x^{t}_{u}$ using as inputs static embeddings of users $x_{u}$ and their corresponding histories' embeddings $h^{t}_{u}$, defined as 
\begin{equation}
h^{t}_{u} = \frac{1}{|\mathbf{h^{t}_{u}}|} \sum_{i \in \mathbf{h^{t}_{u}}} x_{i}~.
\end{equation}
If $\mathbf{h^{t}_{u}}$ is empty, we use $h^{t}_{u}=0$. Respectively, the item block produces dynamic item embeddings $x^{t}_{i}$, and the score of a \textit{(u,i)} couple at a given $t$ is given by the scalar product of their dynamic embeddings. In each block, dynamic embeddings are computed using a network of one-dimensional convolution filters of kernel size $1$ along the embedding dimension axis, considering user and history embeddings as channels (see Fig. \ref{archi}.b). Convolutions were chosen because we empirically found that all architectures performing computations involving both $x_{u, l}$ and $h^{t}_{u, l'}$ with $l \neq l' \in [\![1; d ]\!]$ the $l$-th component of the embedding systematically led to poorer performance than a linear component-wise combination of $x_{u}$ and $h^{t}_{u}$. The chosen convolutions can be seen as a variation of the linear component-wise combination with shared parameters across all components, and allow for network depth.

The network is trained using the \textit{bayesian personalized ranking} (BPR) loss \citep{rendle2009bpr}, a surrogate of the ROC AUC score \citep{manning2010introduction}. It is defined in \citep{rendle2009bpr} as
\begin{equation}
L_{BPR} = - \sum_{(u, i, j) \in D} \text{ln } \sigma (x_{uij})~,
\end{equation}
where $D = \left\{(u,i,j) | i \in I^{+}_{u} \wedge j \in I \backslash I^{+}_{u} \right\}$ with $I^{+}_{u}$ the subset of items that were of interest for user $u$ in the considered dataset, and $x_{uij} = x_{ui} - x_{uj}$, with $x_{ui}$ the score of the $(u,i)$ couple. $\sigma$ is the sigmoid function, defined as $\sigma(x) = 1/(1+e^{-x})$. The underlying idea is to rank items of interest for a given $u$ higher than items of no interest for that $u$, and $D$ corresponds to the set of all such possible pairs for all users appearing in the considered dataset. As $D$ grows exponentially with the number of users and items considered, it is  usual to approximate $L_{BPR}$ with negative sampling \citep{mikolov2013distributed}.

In our proposed methodology, scores become time-dependent. Data samples are therefore not seen as couples, but as triplets $\left(t,u,i\right)$. To enforce user-item symmetry in the sampling strategy, we define 
\begin{align}
D^{t}_{u} &= \left\{(t,u,i,j) | i \in I^{t,+}_{u} \wedge j \in I \backslash \mathbf{h^{t}_{u}} \right\} \\
D^{t}_{i} &= \left\{(t,u,v,i) | u \in U^{t,+}_{i} \wedge v \in U \backslash \mathbf{h^{t}_{i}} \right\} 
\end{align}
with $I^{t,+}_{u}$ the subset of items that had a positive interaction with user $u$ at time $t$, and $U^{t,+}_{i}$ the subset of users that had a positive interaction with item $i$ at time $t$. For a given positive triplet $\left(t,u,i\right)$, we  sample either a corresponding negative one in $D^{t}_{u}$ or $D^{t}_{i}$ with equal probability. Note that considering samples as triplets adds a sampling direction, as a couple that was active at a time $t$ may no longer be active at other times $t+t'$ or $t-t"$, $t', t" > 0$. Such sampling strategies will be more extensively studied in further iterations of this work.

\section{Experiments}

We conduct a series of experiments to understand the behavior of the proposed HCF algorithm on a proprietary database of RFQs on governmental bonds from the G10 countries. This database, which describes every day the RFQs performed by the clients of the bank on G10 bonds, accounts for hundreds of clients and thousands of bonds and ranges from 08/01/2018 to 09/30/2019. On average, we observe tens of thousands of user-item interactions every month. 

We examine here the performance of our proposal in comparison to benchmark algorithms in two experiments. Benchmark algorithms, chosen for their respect of the aims outlined in Section \ref{intro}, are a historical baseline and two matrix factorization algorithms trained using a confidence-weighted euclidean distance \citep{hu2008collaborative} and the BPR objective \citep{rendle2009bpr}, that we respectively call \textit{MF - implicit} and \textit{MF - BPR}. MF - implicit is trained with gradient descent, and we adopt for MF - BPR the symmetrical sampling strategy outlined in Section \ref{methodo}. The historical baseline scores a \textit{(user, item)} couple with their observed number of interactions during the considered training period. 

In this section, the performance of our models is evaluated using mean average precision (mAP) \citep{manning2010introduction}, defined as 
\begin{equation}
\text{mAP} = 1/|Q| * \sum_{q \in Q} \text{AP}(q)~,
\end{equation}
where Q is the set of \textit{queries} to the recommender system, and $\text{AP}(q)$ is the average precision score of a given query $q$. To monitor the performance of our algorithms on both the user- and item-sides, we define two sets of queries over which averaging:
\begin{itemize}
\item \textbf{\textit{User-side queries.}} Queries correspond to the recommendation list formulated every day for all users;
\item \textbf{\textit{Item-side queries.}} Queries correspond to the recommendation list formulated every day for all items.
\end{itemize}
These two query sets lead to user- and item-side mAPs that we summarize with a harmonic mean in a symmetrized mAP score used to monitor all the following experiments, as 
\begin{equation}
\text{mAP}_{sym} = \frac{2 * \text{mAP}_{u} * \text{mAP}_{i}}{\text{mAP}_{u} + \text{mAP}_{i}}~.
\end{equation}
The user- and item-sides equally contribute to the symmetrized score since both sides equally matter in our financial context, as seen in Section \ref{intro}. The mAP scoring perimeter corresponds to the Cartesian product of all the users and items observed in the training period\footnote{This scoring perimeter proved to be the fairest with regard to all considered models, as a model is consequently scored only on what it can score and nothing else. Fixing the perimeter for all algorithms to the Cartesian product of all the users and items observed in the maximal training window and attributing the lowest possible score to otherwise unscorable couples only lowers the mAP scores of candidate algorithms that cannot obtain satisfactory results on such window sizes.}.

\subsection{Evolution of forward performance with training window size}
\label{window}

This experiment aims at showing how benchmark algorithms and our HCF proposal behave with regards to stationarity issues. We already advocated the importance of time in the financial setup --- taking the user side, clients' behavior is \textit{non-stationary}, owing to the non-stationarity of the financial markets themselves but also externalities such as punctual needs for liquidity, regulatory requirements, \ldots~In machine learning, this translates into the fact that the utility of past data decreases with time. However, machine learning and more particularly deep learning algorithms work best when provided with large datasets \citep{lecun2015deep}: there is an apparent trade-off between non-stationarity and the need for more training data. Our goal in this experiment is to show that introducing time in the algorithm helps to reduce this trade-off.

To prove this, we examine the evolution of forward performance as the training window size grows. We split the G10 bonds RFQ dataset into three contiguous parts --- a train part that ranges from up to 08/01/2018 to 07/31/2019 (up to one year), a validation part from 08/01/2019 to 08/30/2019 (one month) and a test part from 09/01/2019 to 09/30/2019 (one month). Validation is kept temporally separated from the training period to avoid signal leakages \citep{de2018advances}, and is taken forward to match business needs. For all the considered algorithms, we train an instance of each on many training window sizes ranging from a week to a year using carefully hand-tuned hyperparameters and early stopping, monitoring validation symmetrized mAP. 

\begin{figure}[h]
    \centerline{\includegraphics[width=0.55\linewidth]{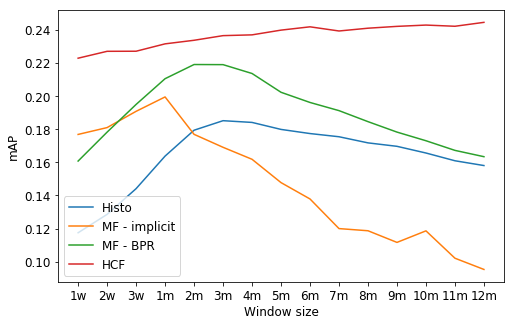}}
    \caption{Evolution of validation symmetrized mAP with training window size. Whereas benchmark algorithms seem to have an optimal training window size, our HCF proposal keeps improving with the training window size.}
    \label{window_size}
\end{figure}

We see in Fig. \ref{window_size} that all benchmark algorithms present a bell-shaped curve. They attain a peak after which their performance only degrades as we feed these models more data, corroborating the non-stationarity vs. amount of data trade-off. On the contrary, HCF only gets better with training data size. Notably, HCF 12m which obtained best validation performance used $n=20$ and blocks with two hidden layers and ReLU activations. 

\begin{figure*}[h]
    \centerline{\includegraphics[width=\linewidth]{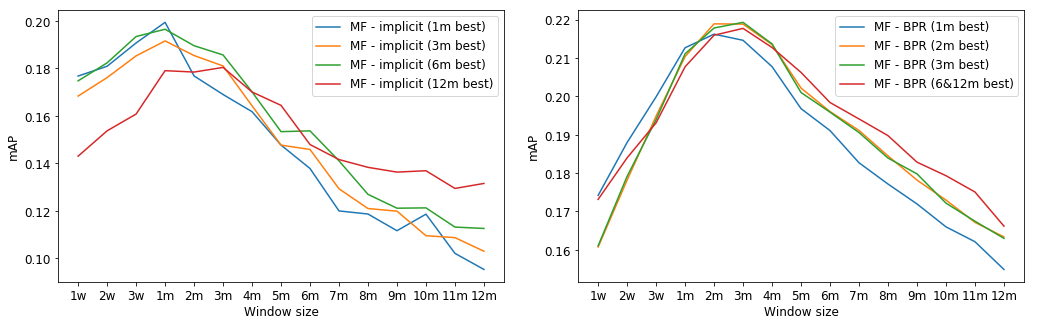}}
    \caption{Evolution of validation symmetrized mAP with training window size. \textit{Left:} Optimal MF - implicit for the 1m, 3m, 6m and 12m windows. A consensus arises around the 1m window. \textit{Right:} Optimal MF - BPR for the 1m, 2m, 3m, 6m and 12m windows. A consensus arises around the 2m-3m windows, slightly favouring the 2m window.}
    \label{bestm_valid}
\end{figure*}

To show that these bell-shaped curves are not an artifact of the hyperparameters chosen in the previous experiment, we conduct a systematic hyperparameter search for multiple training window sizes using a combination of a hundred trials of random search \citep{bergstra2012random} and hand-tuning. The optimal sets of hyperparameters for each considered window size are then used as in the previous experiment to obtain graphs of their validation mAP scores against the training window size. Results are shown in Figure \ref{bestm_valid}. MF - BPR 6m and 12m optimal hyperparameters happened to coincide, and their results are consequently shown on the same graph.

We see here that for both MF algorithms, hyperparameters optimized for many different window sizes seem to agree on optimal window size, respectively around one month and two months, with slight variations around these peaks. Consequently, bell shapes are inherent to these algorithms, which proves their non-stationarity vs. data size trade-off. 

To obtain test performances that are not penalized by the discontinuity of the training and test windows, we retrain all these algorithms with the best hyperparameters and window size found for the validation period on dates directly preceding the test period, using the same number of training epochs as before. The performances of all the considered algorithms are reported in Table \ref{window_results}.

\begin{table}[h]
	\centering
    \caption{Window size study --- symmetrized mAP scores, in percentage.}
    \label{window_results}
    \begin{tabular}{c c c c c}
        \hline
        Algorithm & Window & Valid mAP & Test mAP\\ \hline
        \textbf{Historical} & 3m & 18.51 & 16.86 \\
        \textbf{MF - implicit} & 1m & 19.94 & 19.24 \\
        \textbf{MF - BPR} & 2m & 21.89 & 20.05\\
        \textbf{HCF} & 12m & \textbf{24.43} & \textbf{25.02}\\ \hline
    \end{tabular}
\end{table}

It appears that our HCF algorithm, augmented with the temporal context, obtains better performances on both validation and test periods than the static benchmark algorithms. Consequently, introducing temporal dynamics is essential in the financial setup that we consider.

\subsection{Evolution of forward performance with time}

\begin{figure}[h!]
    \centerline{\includegraphics[width=0.55\linewidth]{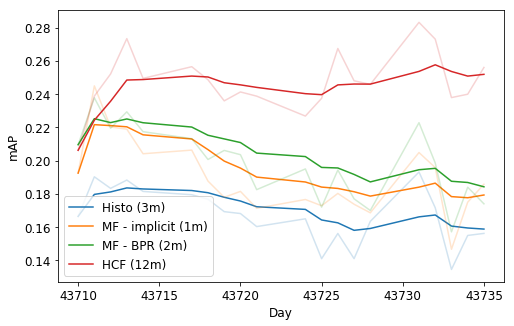}}
    \caption{Daily evolution of symmetrized mAP during the test period. Light shades correspond to the true daily symmetrized mAP values, and dark ones to exponentially weighted averages ($\alpha = 0.2$) of these values.}
    \label{daily_map}
\end{figure}

It follows from the previous experiment that our benchmark algorithms cannot make proper use of large amounts of past data and have to use short training window sizes to obtain good performances compared to historical models. Moreover, the results of these algorithms are less stable in time than HCF results. Fig. \ref{daily_map} visualizes the results reported on Table \ref{window_results} on a daily basis for all the considered algorithms.

We see a downward performance trend for all the benchmark algorithms --- the further away from the training period, the lower the daily symmetrized mAP. On the contrary, HCF has stable results over the whole test period: introducing temporal context through user and item histories hinders the non-stationarity effects on forward performances. 

The results from Section \ref{window} and the observed downward performance trend consequently suggest that benchmark models need frequent retraining to remain relevant regarding future interests. A simple way to improve their results is to retrain these models on a daily basis with a constant, sliding window size $w$ --- predictions for each day of the testing period are made using a model trained on the $w$ previous days. Each model uses here the number of epochs and hyperparameters determined as best on the validation period in the previous experiment. Results of these \textit{sliding} models are shown in Table \ref{sliding_results}, where HCF results corresponds to the previous ones.

\begin{table}[h!]
	\centering
    \caption{Sliding study --- symmetrized mAP scores, expressed in percentage.}
    \label{sliding_results}
    \begin{tabular}{c c c c}
        \hline
        Algorithm & Window & Test mAP \\ \hline
        \textbf{Historical (sliding)} & 3m & 20.32 \\
        \textbf{MF - implicit (sliding)} & 1m & 24.27 \\
        \textbf{MF - BPR (sliding)} & 2m & 24.46 \\
        \textbf{HCF} & 12m & \textbf{25.02} \\ \hline
    \end{tabular}
\end{table}

We see that both MF - implicit and MF - BPR significantly improved their results compared to their static versions from Table \ref{window_results}, but are still below the results of HCF trained on 12 months. Consequently, our HCF proposal is inherently more stable than our benchmark algorithms and captures time in a more efficient manner than their daily retrained versions.

\section{Conclusion}

We introduce a novel HCF algorithm, a time-aware recommender system that uses user and item histories to capture the dynamics of the user-item interactions and that provides dynamic recommendations. In the context of financial G10 bonds RFQ recommendations, we show that for classic matrix factorization algorithms, a trade-off exists between the non-stationarity of users' and items' behaviors and the size of the training datasets. This trade-off is overcome with history-augmented embeddings. Moreover, these embeddings outperform sliding versions of classic matrix factorization algorithms and prove to be more stable predictors of the future interests of the users and items. Further research on the subject will include a more thorough investigation of alternative histories' embeddings formulations and time-aware sampling strategies. Finally, the HCF algorithm could be applied beyond the financial world to tasks where temporal dynamics drive users' behaviors, e.g., music and movie recommendation, where the current mood of a user highly influences her next decisions.

\section*{Acknowledgements}

This work has been conducted under the French CIFRE Ph.D. program, in collaboration between the MICS Laboratory at CentraleSupélec and BNP Paribas CIB Global Markets. We thank Dan Sfedj and Camille Garcin for their work on early versions of the HCF algorithm, and Damien Challet, Sarah Lemler and Julien Dinh for helpful discussions and feedback on drafts of this work.

\newpage
\bibliographystyle{unsrtnat}
\bibliography{references}

\end{document}